\documentclass[conference]{IEEEtran}
\IEEEoverridecommandlockouts
\usepackage{cite}
\usepackage{amsmath,amssymb,amsfonts}
\usepackage{algorithmic}
\usepackage{graphicx}
\usepackage{textcomp}
\usepackage{xcolor}
\def\BibTeX{{\rm B\kern-.05em{\sc i\kern-.025em b}\kern-.08em
    T\kern-.1667em\lower.7ex\hbox{E}\kern-.125emX}}
\begin{document}

\title{Multidomain CT Metal Artifacts Reduction Using Partial Convolution Based Inpainting\\
\thanks{The results have been obtained under the support of the Russian Foundation for Basic Research grant 18-29-26030. The authors thank Skoltech Zhores team for the provided computational resources.}
}
\author{
\IEEEauthorblockN{Artem Pimkin\IEEEauthorrefmark{1},
Alexander Samoylenko\IEEEauthorrefmark{2},
Natalia Antipina\IEEEauthorrefmark{3},
Anna Ovechkina\IEEEauthorrefmark{4},  \\
Andrey Golanov\IEEEauthorrefmark{3},
Alexandra Dalechina\IEEEauthorrefmark{5} and
Mikhail Belyaev\IEEEauthorrefmark{1}}
\IEEEauthorblockA{\IEEEauthorrefmark{1}Skolkovo Institute of Science and Technology, Moscow, Russia\\
Email: \{a.pimkin, m.belyaev\}@skoltech.ru}
\IEEEauthorblockA{\IEEEauthorrefmark{2}Kharkevich Institute for Information Transmission Problems, \\ Moscow Institute of Physics and Technology, Moscow, Russia\\
Email: samojlenko.ai@phystech.edu}
\IEEEauthorblockA{\IEEEauthorrefmark{3}N. N. Burdenko National Medical Research Center of Neurosurgery, Moscow, Russia \\
Email: \{nantipina, golanov\}@nsi.ru}
\IEEEauthorblockA{\IEEEauthorrefmark{4}Lomonosov Moscow State University, Moscow, Russia \\
Email: av.ovechkina@physics.msu.ru}
\IEEEauthorblockA{\IEEEauthorrefmark{5}\textit{Moscow Gamma Knife Center}, Moscow, Russia \\
Email: adalechina@nsi.ru}
}

\maketitle

\begin{abstract}
Recent Metal Artifacts Reduction (MAR) methods for Computed Tomography are often based on image-to-image convolutional neural networks for adjustment of corrupted sinograms or images themselves. In this paper, we are exploring the capabilities of a multidomain method, which consists of both sinogram correction (projection domain step) and restored image correction (image-domain step). We formulate the first step problem directly as sinogram inpainting, which allows us to use methods of this specific field, such as partial convolutions. Moreover, we propose a synthetic data generation pipeline to avoid problems with overfitting to metal shapes set and an artifacts formation technique. The proposed method achieves state-of-the-art ($-75\%$ MSE) improvement in comparison with a classic benchmark - Li-MAR.
\end{abstract}


\section{Introduction}
%Computer Tomography is based on the measurement of the attenuation of X-Rays being taken from a range of different angles.
\subsection{General}
Computed Tomography (CT) is a commonly used imaging method in disease diagnosis and treatment planning. In particular, dose distributions in radiation therapy are calculated as the solution of a forward problem; CT images determine the electron density of the irradiated tissues and patient-specific anatomy. High-density objects (e.g., containing metal) may occur in the area of interest and strongly affect the attenuation of X-Rays that may lead to distortion of the final image reconstructed from an inconsistent sinogram \cite{boas2012ct}. These image artifacts could have a significant impact on the dose calculation accuracy and reduce the visibility of organs and structures close to the metal objects \cite{giantsoudi2017metal}, \cite{kovacs2018metal}.

\subsection{CT artifacts in Brain Radiosurgery}
Though CT artifacts may affect the quality of radiation therapy in various clinical scenarios, we focus on cerebral arteriovenous malformations (AVMs), which are the focal conglomerations of the pathological vessels in the brain.
The primary goal of the AVM management is to prevent the risk of intracranial hemorrhage \cite{lawton2015brain}. AVMs treatment options include microsurgery, radiosurgery, embolization or combination of these modalities \cite{derdeyn2017management}.
Within our study, we analyzed CT scans of the patients with AVMs who underwent prior embolization to block blood circulation within pathological vessels before radiosurgery. 

\begin{figure}[htb]
\centering
\includegraphics[scale=0.34]{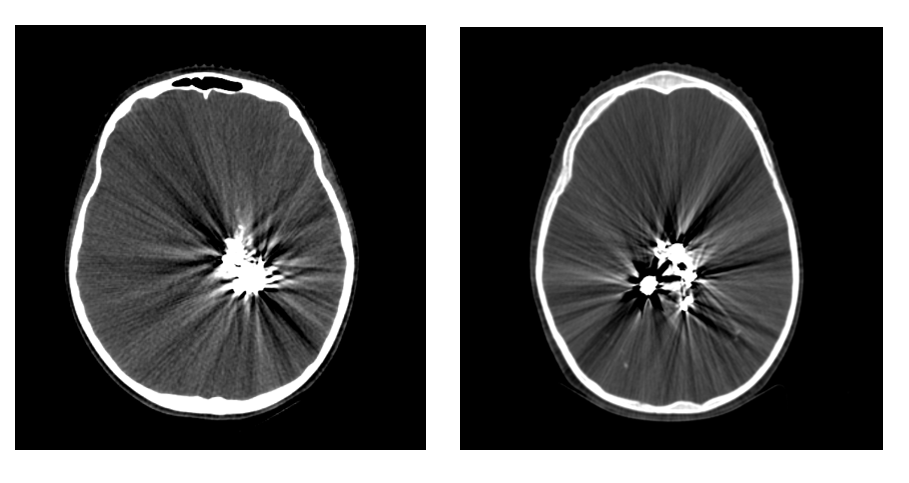}
\caption{Liquid (left) and solid (right) embolization of the AVMs with notable artifacts on the CT images due to the metal components of the embolic materials.}
\label{fig:embolization_examples}
\end{figure}

Embolic agents injected during the first stage of this treatment can be divided into solids (for example, metal coils) and liquids. Metal components of the embolic agents, as well as coils' metal body, cause severe artifacts on CT scans, producing both bright regions of high absorption and dark regions of low absorption.
Several studies have proposed that the presence of these high-density objects induces beam hardening artifacts on CT scans and might distort the dose calculation accuracy of the radiosurgery planning \cite{shtraus2010radiosurgical}. Figure \ref{fig:embolization_examples} shows CT scans of two patient with AVMs after embolization (see more technical details on types of embolization in Section \ref{ssec:data}). 

The artifacts need to be identified during treatment planning and somehow fixed to avoid possible dose calculations errors \cite{roberts2012dosimetric}. In current clinical practice, this procedure is usually manual: doctors segment primary tissues like brain or skull and then replace electron densities of corrupted areas by the electron densities appropriate to the tissue. This time-consuming and a rather rough operation must be repeated for each affected slice of the CT image (usually 50-100 2D images for one patient).

%It is commonly used as a diagnostic tool, but it is also applied, for example, in radiotherapy for calculation of doses distribution. High density objects (e.g. containing metal) may occur in the area of interest and strongly affect the attenuation of X-Rays that may lead to distortion of the final image reconstructed from inconsistent sinogram \cite{boas2012ct}. That led to the development of different Metal Artifacts Reduction (MAR) methods during the last 40 years \cite{gjesteby2016metal}.

%These image artifacts interfere the process of doses distribution calculation and commonly make it impossible. 

\subsection{Problem}
Thus we formulate a Metal Artifacts Reduction problem as the following: given an input 3D tensor that represents a CT scan corrupted by the presence of high-density objects, generate a corresponding CT image with suppressed artifacts. 
Due to the high importance of the MAR, a variety of different methods has been developed during the last 40 years; see an overview \cite{gjesteby2016metal}. It's important to note that in addition to usual image space, each CT slice can be naturally represented as a sinogram, which mathematically is the direct Radon transform of the same slice in the image space. Indeed, the majority of the proposed solutions can be divided into two large groups based on the domain of the input data:
\begin{enumerate}
    \item Algorithms consisting of removal of the high-density area from sinogram with further reconstruction based on the uncorrupted parts, also known as projection-based, e.g., classical Li-MAR which consists of linear interpolation for missing data within the metal trace. \cite{kalender1987reduction}. Nowadays, this problem may be solved using image-to-image deep convolutional networks (e.g., \cite{park2017sinogram}).     
    \item Image-based solutions that use image-to-image networks and reduce artifacts directly on the pixel data(e.g., \cite{park2017machine}).
\end{enumerate}

\subsection{Contribution}
\textit{First}, we propose a deep learning-based method that combines both approaches described above: it consists of two models that process the image representation in two domains. The first inpainting model is responsible for the removal of the distorted metal trace from the sinogram. The second refining model corrects the residual artifacts after image restoration. In this work, we have successfully verified the following statements: 
\begin{enumerate}
    \item Refining model improves the quality of the result since even minor inconsistencies in the sinogram may lead to significant artifacts on the restored image.     
    \item Sinogram adjustment via the inpainting model may simplify the problem for the direct image-to-image refining model.
\end{enumerate}
And while the idea of the dual-domain method already appeared in some works (e.g. \cite{Dudo}), we formulate the problem of restoration of the corrupted area of sinogram directly as image inpainting. It allows us to use state-of-the-art approaches for this step, i.e., partial convolution-based neural network proven to outperform classic fully-convolutional end-to-end approach (method was introduced in \cite{Liu2018ImageIF} for inpainting of irregular holes).

\textit{Second}, we propose a new pipeline for random generation of high-density objects that allows us to train the networks in a self-supervised setting due to the absence of (artifact-free CT, CT with artifacts) pairs. Our method doesn't rely on simple artifacts generation in the image domain. This is an important part of the proposed method since it allows us to avoid two major problems of the common approaches of the synthetic datasets for artifacts reduction: (1) mismatch between simple generation algorithms and the diversity of the shapes on real CT scans of AVMs patients and (2) possible overfitting to the structure of the simulated artifacts (see more details in Section \ref{sec:synth_data}).  

We compared the performance of the proposed model with FCN-MAR\cite{ghani2018deep}, Deep-MAR\cite{ghani2019fast} and CNN-MAR\cite{zhang2018convolutional} as one of the most popular and recent deep learning based methods of metal artifact reduction. The results show the superior quality of our method.

\begin{figure}[htb]
%\centering
\includegraphics[scale=0.34]{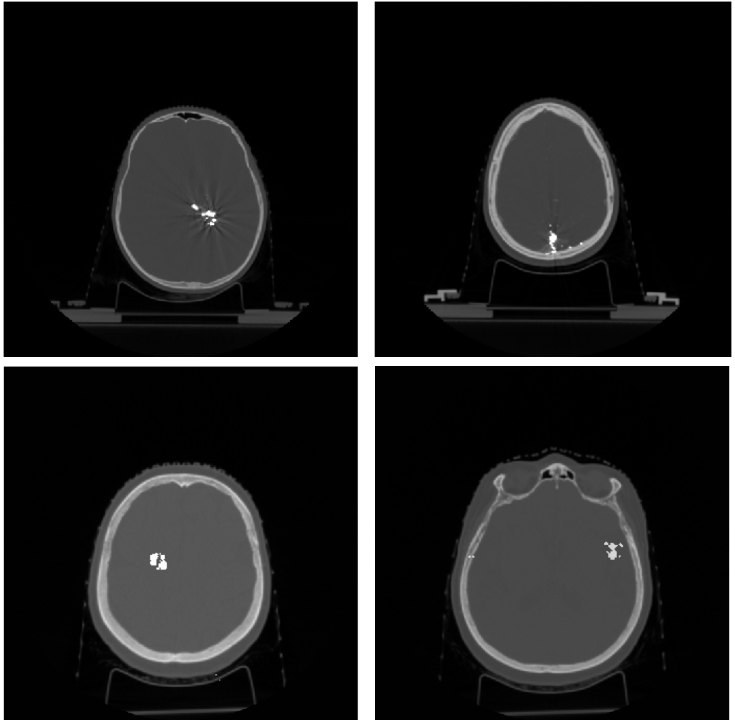}
\caption{Example of real shapes of high density objects (above) and shapes generated via proposed pipeline (below).}
\label{fig:gen_example}
\end{figure}

\begin{figure*}[htb]
\centering
\includegraphics[scale=0.34]{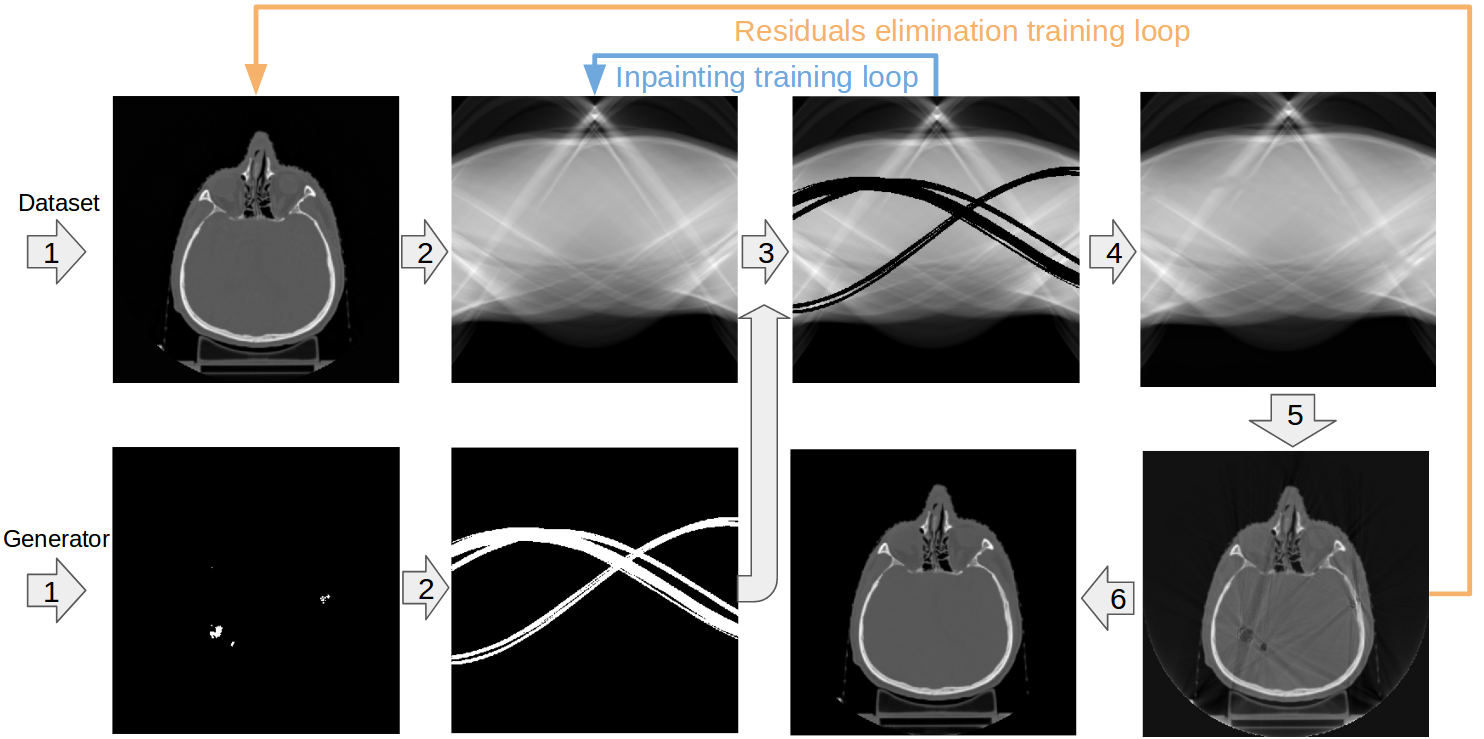}
\caption{Overall training pipeline. Step 1: random generation of high-density objects.  Step 2: sinogram calculation for (a) input CT image an (b) generated objects. Step 3: removing parts of sinogram (a) using sinogram (b) as a binary mask. Step 4: the output of the inpainting model (the blue arrow highlights training input-output pairs). Step 5: inverse Radon transform of (4).  Step 6: suppression of the residual artifacts by the image-to-image network (the orange arrow represents training data).}
\label{fig:training}

\end{figure*}

\section{Proposed solution}
\label{sec:typestyle}

The main idea of the proposed solution is to combine both image-based and projection-based approaches. The first step is the removal of the metal trace from the sinogram with the restoration of deleted areas (projection domain step). The second step is the elimination of residual artifacts from the restored image (image domain step). A more accurate formulation of the pipeline is further in Section \ref{ssec:structure}. 
% Here you need to talk about the main idea: we split the solution in two steps: high density objects removal (1), and artifacts elimination. Написал.

% The you split the section in two subsection and discuss each step. Present your motivation: why inverse Radon, why impainting, why you need the second net. Не согласен, мотивация дана во введении, дублировать ее нет смысла (и места), равно как и дробить повествование на два блока. 

\subsection{Overall structure}
\label{ssec:structure}

Firstly, it is important to mention that we implemented slice by slice pipeline due to avoidance of spatial data inconsistencies, e.g., different spacing between slices, which is common for medical imaging. Thus, we propose the following algorithm structure for each of the corrupted slices:
\begin{enumerate}
    \item Cut a mask of the high-density object using threshold (since CT voxel intensities represent its density). It's a common method to detect such objects (e.g \cite{ghani2019fast}).
    \item Use Radon transform to obtain sinograms (a) for the corrupted image itself and (b) for the mask of the found high-density objects. 
    \item Remove these parts of sinogram (a), which corresponds to the sinogram (b), i.e., affected by the high-density objects.
    \item Restore cropped sinogram area using an inpainting network (see details below). Ideally, it has to match the sinogram of the image with the absence of both high-density objects and artifacts caused by it. 
    \item Calculate the image from the modified sinogram using the inverse Radon transform.
    \item Adjust the image to remove residual artifacts using the second, refining networks (see details below).
    \item Add the high-density object to the final image using the mask from step 1.
\end{enumerate}

At steps 4 and 6 we use convolutional neural networks as described in Section \ref{ssec:architectures}. 
To measure the effectiveness of the proposed algorithm in comparison with only sinogram inpainting and direct image to image model, we also trained the same pipeline without step 4 and step 6, respectively. More details on the pipeline and training loops can be found in Figure \ref{fig:training}.

\subsection{Synthetic data}
\label{sec:synth_data}
Due to the absence of the paired images, we decided to create an algorithm to generate realistic high-density objects and train networks in self-supervised settings. We used CT scans of the patients with no such high-density objects and related artifacts to create a synthetic dataset.  We built a random 3D-shape generator such that generated objects had a similar structure as real high-density objects on head CT scans. Indeed, as we mentioned in the introduction, we aimed to solve two common issues with algorithms for the generation of the synthetic datasets for artifacts reduction: \begin{enumerate}
    \item \underline{Shapes diversity on a real scan data.} Common method of high-density shapes generation is creating a dataset of objects from the real scans (e.g. \cite{Dudo}, \cite{ghani2018deep}), which is not so applicable to the field of radiosurgery due to wide range of the shapes of AVMs after endovascular embolization. That is why for this specific case of metal objects appearing in radiosurgery practice, we suggest using a generator of a structurally close to the real, but still random and diverse shapes, which leads to the increased robustness of the model.        
    \item \underline{Artifacts characteristics overfitting.} MAR algorithms that operate in the image domain are indirectly relying on the structure of the artifacts (e.g., \cite{zhang2018convolutional}). However, as those artifacts are simulated, it may introduce a systematic bias and a subsequent drop in models' performance on real artifacts. Thus, we suggest using only the binary mask of high-density objects with no utilization of an image with primary artifacts.   
\end{enumerate}

To achieve this, we propose the following algorithm:
\begin{enumerate}
    \item Select a volumetric range to place objects randomly (uniformly from 0 \% up to 10\% of the linear size of the image). 
    \item Put a random number (uniformly from 1 to 25) of geometrical structures (ball, octahedron, parallelepiped) of linear size up to 10 pixels into this volume.
    \item Merge these random structures using the morphological closing operation to obtain a randomly shaped object or a small set of them.
    \item Put into the volume up to 30 geometrical structures of small size (from 1 to 3 pixels) to obtain an object with outliers. 
    \item Select a position of the obtained object randomly (via uniform 2D distribution) so that the overlap between the mask and the brain is $\geq 95\%$.
    \item Repeat the process up to 10 times to obtain a scan with multiple objects
\end{enumerate}
Figure \ref{fig:gen_example} shows examples of real and generated objects.
Using that algorithm for each CT image, we created 30 differently distributed object masks with an approximate depth of 90 slices per sample on average. 

\subsection{Network architectures}
\label{ssec:architectures}
For deep learning problems formulated above, we used the following architectures:
\begin{enumerate}
    \item \textbf{UNet\cite{unet}} - a well known fully-convolutional network with additional connections between encoder and decoder. As it was originally proposed for small medical datasets, Unet is also performed well in the task of metal artifacts reduction on sinograms (e.g. \cite{park2017sinogram}) as well as in the direct image to image artifacts removal process (e.g. \cite{park2017machine}). We use this architecture for step 6 of our algorithm (reduction of the residual artifacts on the images).
       
    \item \textbf{UNet with partial convolutions} - this architecture differs from the UNet mentioned above with convolutions that were replaced by partial convolutions \cite{Liu2018ImageIF} which are masked and renormalized to be conditioned only on the pixels that are not masked. In this work, we formulate the problem of sinogram correction directly as inpainting, which allows us to integrate this state-of-the-art method to our pipeline directly. 
\end{enumerate}

\section{Experimental setup}
\label{ssec:setup}
\subsection{Data}
\label{ssec:data}
 Our data consisted of two datasets from Radiosurgery Department at the Burdenko Neurosurgical institute. To obtain CT scans without artifacts, we selected 163 CT scans of the patients treated with radiosurgery for trigeminal neuralgia (TN). All these patients underwent computed tomography as a part of the planning process. For the acquisition of the CT images, GE Optima 580 was used. The scanning parameters were as follows: 120 kVp, 350 mA, FOV 300 mm, slice thicknesses 1,5 mm. 

Each scan is represented by a 3D tensor of shape $512\times 512\times N$, where N is the number of axial slices varying from 130 to 210. The data was divided into three parts patient wisely: a training set for the first model, a training set for the second model (approx. 10000 $512\times 512$ images for both training sets), and an independent testing set (approx 9000 $512\times 512$ images).
 
Also, we selected 47 CT scans of the patients with AVMs that were treated with both radiosurgery and prior endovascular embolization or surgery between 2014 and 2019. Embolic agents can be divided into solids (for example, metal coils) and liquids. Two commonly used liquid materials for cerebral AVM embolization are Onyx (ethylene-vinyl alcohol copolymer (EVOH))and n-BCA (n butyl cyanoacrylate) \cite{roberts2012dosimetric}. The n - BCA agent is a fast - polymerizing liquid adhesive. It includes iodine (Z = 53) as a visualization material during the injection. Onyx contains suspended micronized tantalum (Z = 73) powder. 
As we discussed, metal components of the embolic agents cause severe artifacts on CT scans, producing both bright regions of high absorption and dark regions of low absorption. This dataset was used in two ways
\begin{enumerate}
    \item for the qualitative validation of the proposed solution by experts;
    \item study the shapes and sizes of real metal-containing objects to design our method for synthetic data generation. 
\end{enumerate}

\subsection{Preprocessing}
\label{ssec:preprocessing}
All input images for both sinogram inpainting and restored CT slice residuals reduction were linearly normalized to fit into $[0, 1]$ window. Such a simple preprocessing was used to maintain the physical sense of voxel intensities.

\begin{figure*}[]
    \centering
    \includegraphics[scale=0.31]{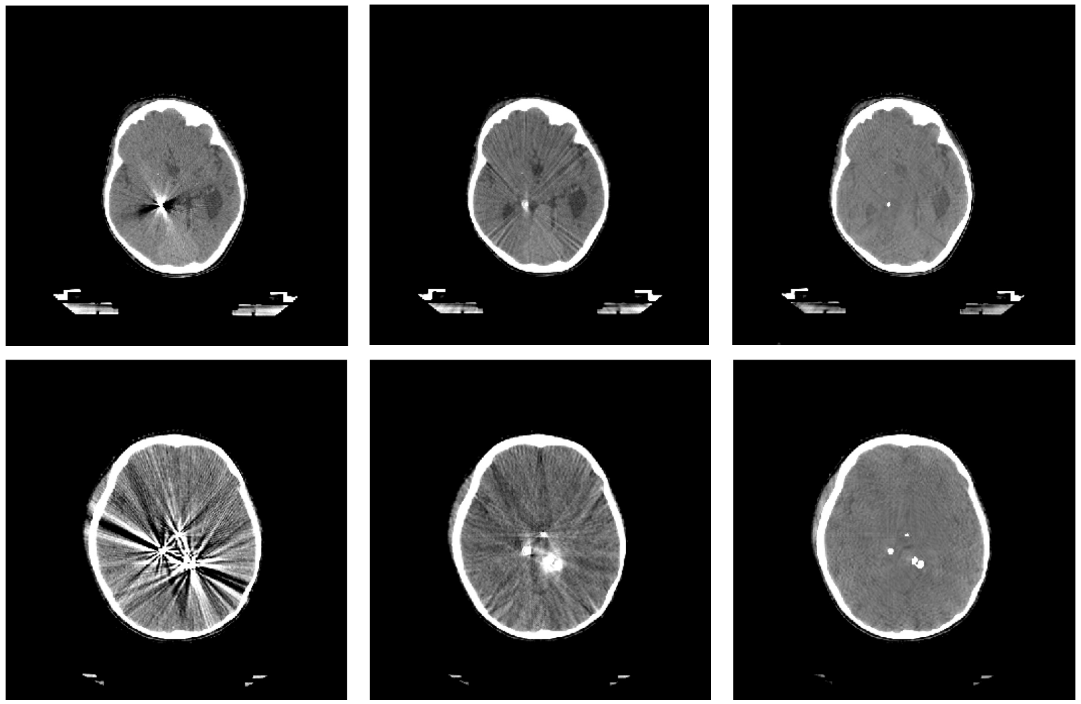}
    \caption{Examples of processing real corrupted CT scan with intensities in brain window (40+-80 HU). From left to right: original image, Li-MAR output, the output of the proposed solution. Each row corresponds to a different patient.}
    \label{fig:work_example}
\end{figure*}

\subsection{Metrics}
\label{ssec:metrics}
We used a standard MAE (Mean Absolute Error), MSE (Mean Squared Error), and SSIM (Structural Similarity) metrics to measure the quality of artifact reduction on the CT scans. In all cases, we compared the original image with the artificially corrupted and then restored one.

We validated our models in two ways:
\begin{enumerate}
    \item We compared our multidomain method with image-domain only and sinogram-domain only networks, as we described in Sections \ref{ssec:structure} and \ref{ssec:architectures}. 
    \item For comparison with other recent convolutional network based methods, we've implemented classic Li-MAR and measured relative increase of performance in terms of MSE. Due to the absence of the public MAR datasets and a large number of possible pitfalls in reimplementing architectures from other works, we examined the quality of our approach using the relative difference between model and Li-MAR performances as a metric. This is a common choice for the field: relative improvement of Li-MAR performance is often reported across papers (e.g., \cite{ghani2018deep}, \cite{ghani2019fast} and \cite{zhang2018convolutional}), which allows us to compare complex methods despite the absence of the public dataset.
\end{enumerate}

\subsection{Training}
\label{ssec:training}
For our full pipeline and approaches comparison, we end up with training 3 models: sinogram inpainting (UNet with partial convolutions), residual artifacts elimination (UNet) and image-to-image model for the pipeline without inpainting step (Unet).
%from the slice restored from sinogram with the cropped area to the slice with reduced artifacts (UNet).
All of the models mentioned above were trained using Adam optimizer with the initial learning rate of $5\cdot10^{-3}$. 

Sinogram inpainting UNet with partial convolutions was trained for 500 epochs with the multiplication of the learning rate by $0.5$ on each of the following epochs: 100, 200, 300 and multiplication by $0.1$ on 400, 450, 475 and 490 epochs respectively.  Models with plain UNet architecture were both trained for 200 epochs with $0.5$ learning rate multiplication on 100, 150 and 175 epochs.

All models were trained on Zhores supercomputer \cite{zacharov2019zhores} using PyTorch framework and DPipe\footnote{https://github.com/neuro-ml/deep\_pipe} for configurations management and experiments setup.

\section{Results}
\label{sec:results}

\begin{figure}[]
\centering
\includegraphics[scale=0.48]{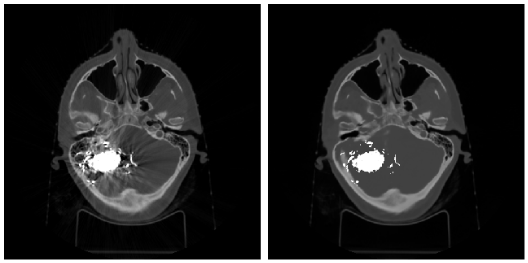}
\caption{An example of a CT scan with large metal-containing object causing significant distortion (left) and the output of the proposed solution (right).}
\label{fig:bigobj}

\end{figure}

As we mentioned above, the classical benchmark for metal artifacts refuction methods is Li-MAR.
Figure \ref{fig:work_example} shows examples of our algorithm and Li-MAR work on real CT scans of the patients with high-density objects that cause artifacts on the images. It is visible in the brain window that the model restores distinguishable brain structures well. 
Also, it is important to mention that the qualitative expert analysis using CT studies of patients with AVMs demonstrates that the method achieves impressive results on scans with large metal regions, which caused significant image distortion. An example is shown in Figure \ref{fig:bigobj}.

\begin{table}[htb]
\centering
\caption{Comparison between one step methods, Li-MAR and two step method in terms of MAE, MSE and SSIM between the final output and test set. We used original pixel intensities in Hounsfield units to calculate these metrics.}
\resizebox{\columnwidth}{!}{%
\begin{tabular}{|l|l|l|l|}
\hline
Algorithm & MAE (HU) & MSE (HU$^2$)& SSIM\\ \hline
Li-MAR                & 26.6 & 3282 & 0.94 \\ \hline
Inpainting-only       & 22.0 & 10170 &0.94  \\ \hline
Image-to-image only   & 14.5  & 2064 & 0.97  \\ \hline
Proposed algorithm    & \textbf{9.6}  & \textbf{831}& \textbf{0.99}   \\ \hline
\end{tabular}
}
\label{tbl:res}
\end{table}

All the test metrics of inpainting-only, image-to-image only, Li-MAR as a classic benchmark, and the proposed method are represented in Table \ref{tbl:res}. 
Here we can see a significant ($~56\%$) decrease of MAE between the inpainting-only method and the proposed solution. Thus we can conclude that the image-to-image network can successfully remove residual artifacts and increase the quality of the joint model. We obtained quite large MSE and relatively small MAE for the inpainting model due to the inhomogeneity of its errors caused by sinogram inconsistency. On the other hand, we can see a decrease of MAE of $~34\%$ between image-to-image only method and our two-step approach. This confirms our hypothesis that sinogram inpainting as preprocessing simplifies the task for the image-to-image model. Moreover, the proposed solution significantly outperforms Li-MAR by $64\%$ in terms of MAE.

Table \ref{tbl:comp} shows a relative drop of MSE reported in different papers in comparison with the reported Li-MAR score and provided us an understanding of the relatively good performance of the proposed model. For comparison, we took three recent deep-learning-based methods operating in one or both domains:
\begin{itemize}
    \item \textbf{FCN-MAR\cite{ghani2018deep}} - method based on sinogram completion via convolutional network trained on fully synthetic data.
    \item \textbf{Deep-MAR\cite{ghani2019fast}} - conditional generative adversarial network (cGAN\cite{mirza2014conditional}) based method for sinogram completion trained on fully synthetic dataset with launch on real data via transfer learning.
    \item \textbf{CNN-MAR\cite{zhang2018convolutional}} - dual-domain method consisting of processing scans in image domain with different simple methods (e.g. Li-MAR) for usage as input for CNN. The output of the neural network is used to differentiate tissues, generate a prior image and, therefore, sinogram to replace metal trace in the projection domain.
\end{itemize}

\begin{table}[h]
\centering
\caption{Relative drop compared to Li-MAR performance.  
}
%\resizebox{\columnwidth}{!}{%
\begin{tabular}{|c|c|}
\hline
Algorithm & MSE drop \\ \hline
Li-MAR \cite{kalender1987reduction}    &   0.0$\%$ \\ \hline
FCN-MAR \cite{ghani2018deep}           & -52.2$\%$ \\ \hline
Deep-MAR \cite{ghani2019fast}          & -58.5$\%$ \\ \hline
CNN-MAR \cite{zhang2018convolutional}  & -71.1$\%$ \\ \hline
Proposed algorithm    & \textbf{-75.5$\%$}\\ \hline
\end{tabular}
%}

\label{tbl:comp}
\end{table}

\section{Discussion}

In this paper, we have presented a state-of-the-art multidomain deep learning procedure that consists of two main parts. Firstly, it is a metal shape generator that allows to transfer resulting model directly to the real data without using finetuning, transfer learning or domain adaptation techniques. Secondly, the method combines two CNN models: sinogram completion via state-of-the-art partial convolutions based CNN and additional adjustment after inverse Radon transform. The results of our experiments show the effectiveness of the proposed approach compared to one-step algorithms. Moreover, qualitative and quantitative analysis shows an application potential of the proposed method and its applicability, even in difficult cases. Thus, the next step of our research is going to be a clinical validation of the dose calculations for radiosurgery planning.

\bibliographystyle{IEEEbib}
\bibliography{strings,refs}

\end{document}